\documentclass{article}

\usepackage{geometry}
\usepackage{amsmath, amsthm, amsfonts, bm, amssymb, enumitem, tabularx}
\usepackage{amsfonts, graphicx, grffile,fullpage, float, latexsym, mathtools, bbm}
\usepackage{hyperref}
\hypersetup{
    colorlinks=true, %set true if you want colored links
    citecolor=black,
     linkcolor=blue,
     urlcolor=blue,
    linktoc=all,     %set to all if you want both sections and subsections linked
    linkcolor=black,  %choose some color if you want links to stand out
}
\usepackage{cleveref}
\usepackage{multicol}
\usepackage{tikz}
\usepackage{subcaption}
\usepackage{algpseudocode}
\usepackage{algorithm}

\algnewcommand{\Inputs}[1]{%
  \State \textbf{Inputs:}\hspace*{\algorithmicindent}\parbox[t]{.8\linewidth}{\raggedright #1}
}
\algnewcommand{\Initialize}[1]{%
  \State \textbf{Initialize:}\hspace*{\algorithmicindent}\parbox[t]{.8\linewidth}{\raggedright #1}
}

\newcommand{\rev}[1]{\textcolor{red}{#1}}

\usepackage[sort, numbers, compress]{natbib}

\usepackage{Shorthands}

\newcommand{\change}[1]{\textcolor{red}{#1}}

\begin{document}

\title{ADMM-MCBF-LCA: A Layered Control Architecture \\ for Safe Real-Time Navigation}
\author{Anusha Srikanthan*, Yifan Xue*, Vijay Kumar, Nikolai Matni, Nadia Figueroa
\thanks{*-Equal Contribution}
\thanks{We gratefully acknowledge the support of NSF Grant CCR-2112665, NSF awards CPS-2038873, SLES-2331880, AFOSR Award FA9550-24-1-0102 and NSF CAREER award ECCS-2045834. for this research. %TILOS (AI, multirobot, autonomy, mapping, trajectory generation)
}
\thanks{A. Srikanthan, Y. Xue, V. Kumar, N. Matni, and N. Figueroa are with the School of Engineering and Applied Science, University of Pennsylvania, Philadelphia, USA (e-mail:\{sanusha, yifanxue, kumar, nmatni, nadiafig\}@seas.upenn.edu). }}

\maketitle

\begin{abstract}
    We consider the problem of safe real-time navigation of a robot in \rev{a dynamic environment with moving obstacles of arbitrary geometries} and input saturation constraints. We assume that the robot detects nearby obstacle boundaries with a short-range sensor \rev{and that this detection is error-free}. This problem presents three main challenges: i) input constraints, ii) safety, and iii) real-time computation. To tackle all three challenges, we present a layered control architecture (LCA) consisting of an offline path library generation layer, and an online path selection and safety layer. To overcome the limitations of reactive methods, our offline path library consists of feasible controllers, feedback gains, and reference trajectories. To handle computational burden and safety, we solve online path selection and generate safe inputs that run at $100$ Hz. \rev{Through simulations on Gazebo and Fetch hardware in an indoor environment}, we evaluate our approach against baselines that are layered, end-to-end, or reactive. Our experiments demonstrate that among all algorithms, only our proposed LCA is able to complete tasks such as reaching a goal, safely. When comparing metrics such as safety, input error, and success rate, we show that our approach generates safe and feasible inputs throughout the robot execution.
\end{abstract}

\section{Introduction}

Historically, motion planning algorithms~\cite{canny1988complexity} decouple the planning problem from feedback control design. The challenge today is to integrate these approaches to ensure safe robot behavior. To this end, layered control architectures (LCAs) offer a design methodology for robot navigation using hierarchical layers balancing constraint complexity and trajectory horizons: see the recent survey~\cite{matni2024towards} which highlights the success of LCAs across robotics and other application areas.  Many works~\cite{rosolia2020multi, grandia2021multi, rosolia2022unified, matni2024towards} employ LCAs where higher layers solve model predictive control (MPC) over long horizons at low frequencies, while lower layers handle trajectory tracking at higher frequencies. However, integrating these layers is challenging due to mismatched time scales, real-time computational requirements, and task completion objectives. Real-time solutions~\cite{stolle2006policies, majumdar2017funnel} often rely on offline libraries of 
% infeasible 
reference trajectories or reachable sets, combined with online planning, but these components assume partial or complete knowledge of the environment and can be difficult to coordinate effectively. In contrast, we assume that obstacles within a local field of view are visible to the robot, and aim to ensure safety through reactive control.

\begin{figure}[t]
    \centering
    \begin{subfigure}{0.597\linewidth}
         \includegraphics[width=\textwidth]{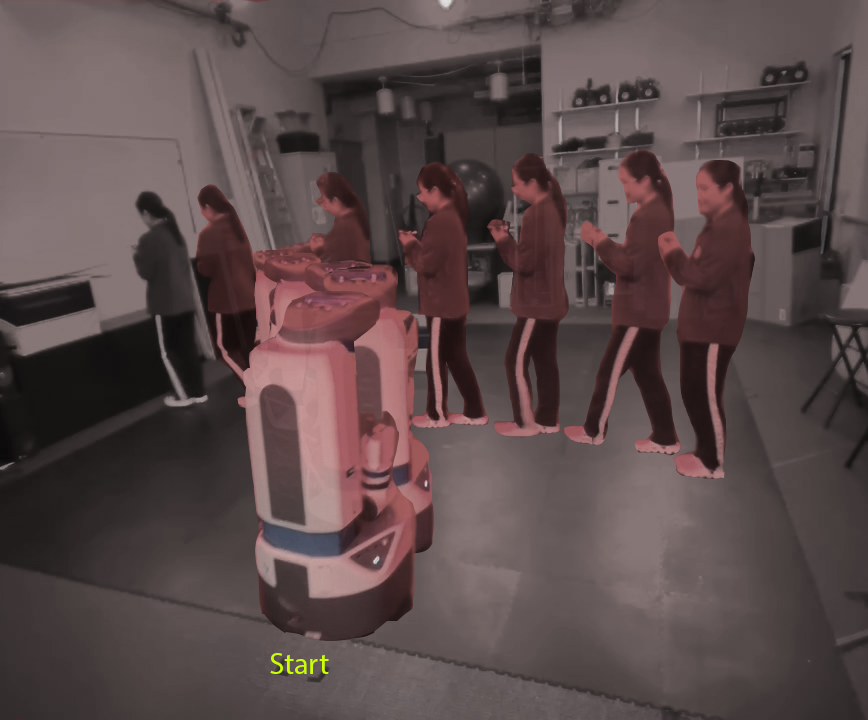}
    \caption{Safe Real-Time Navigation\label{fig:video}}
\end{subfigure}\hspace{2pt}\begin{subfigure}{0.3899\linewidth}
         \includegraphics[trim={11cm 2cm 17cm 10cm},clip,width=\linewidth]{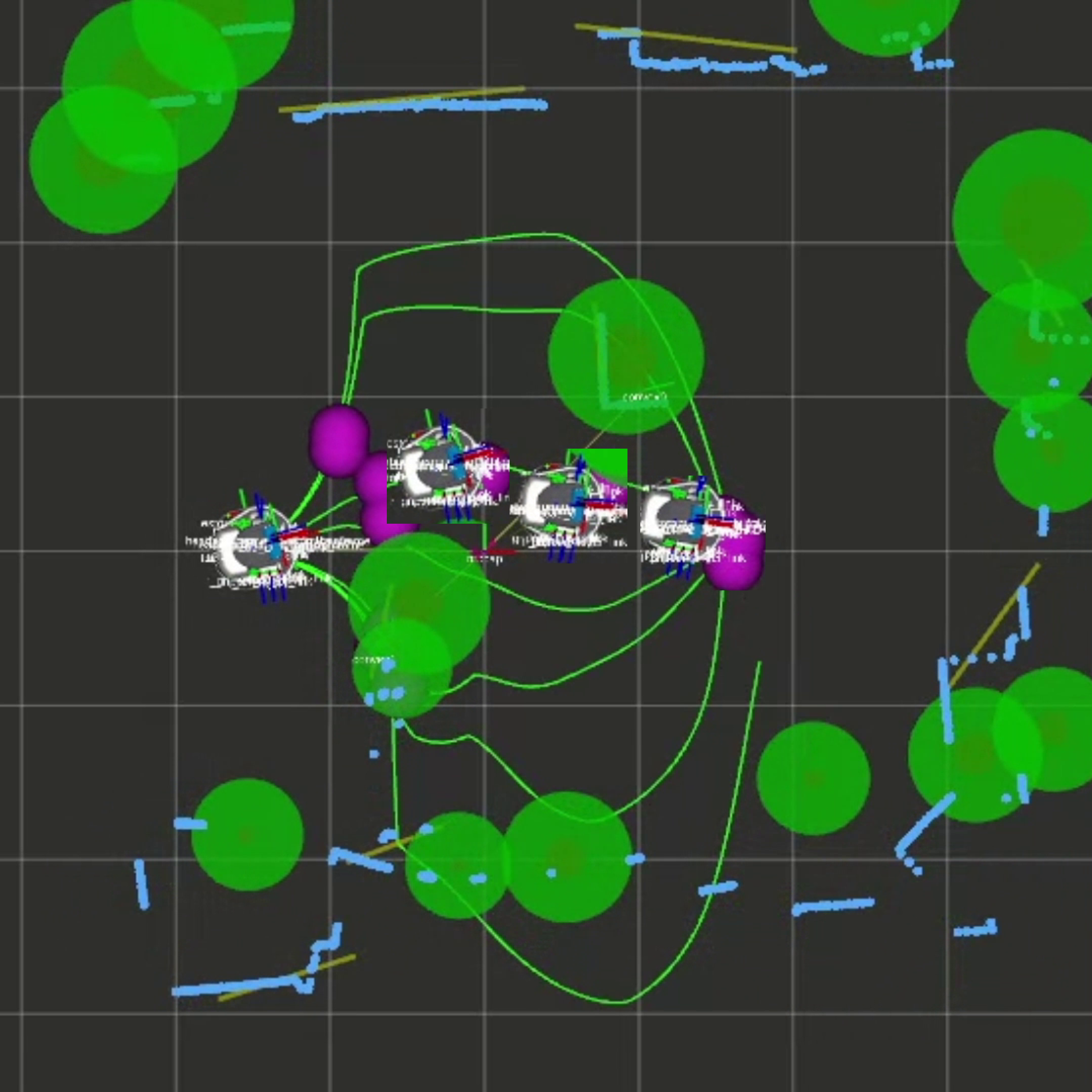}
     \caption{ADMM-MCBF \label{fig:rviz}}
    \end{subfigure}
    \caption{We show a Fetch robot successfully reaching a goal in a dynamic environment running our proposed algorithm to avoid a human walking around and other static obstacles. Blue and green shades in \autoref{fig:rviz} are obstacles, while the green trajectories are ADMM-generated feasible paths. }   
    \label{fig:experiment}
\end{figure}

Although reactive control strategies for safety-critical robot navigation~\cite{hsu2023safety}, such as artificial potential fields~\cite{apf1989real}, navigation functions~\cite{loizou2003closed, loizou2007mixed}, dynamical system (DS) based modulation~\cite{LukesDS,DSTEXTBOOK} and control barrier functions (CBFs)~\cite{ames2019control,notomista2021safety,singletary2021comparative}, are computationally effective; they are limited by their greedy policies for long-horizon planning for control-affine input-constrained systems.

Such policies struggle to escape local minima (saddle points) in cluttered non-convex environments that are not ball \cite{notomista2021safety} or star worlds \cite{starworlds} resulting in sub-optimal trajectories and failed mission objectives. 
% This leads to suboptimal trajectories and failed task completion in realistic dynamic environments. 
Further, designing reactive control strategies for nonholonomic vehicles, such as differential drive robots, adds the challenge of contradicting input and safety constraints alongside the use of simplified dynamics models~\cite{planning}. Reactive methods largely rely on a nominal global trajectory encoding task objectives for mission completion. %Task completion for reactive methods largely is governed by the quality of the nominal global trajectory. 
Unlike planning methods that design reference trajectories, we generate path libraries consisting of nominal controllers, feedback gain matrices and reference trajectories. This allows us to relax input constraints at the reactive layer, as we will show, drastically reducing solver infeasibility that plague reactive controllers. 

Therefore, in this paper, we offer an LCA that generates an offline library of feasible paths for control-affine input-constrained systems \rev{also allowing for smooth switching between paths}, used by an online path selector that tracks the closest global trajectory with a reactive control strategy ensuring safety in real-time. While one could use any of the aforementioned reactive control strategies, in our proposed LCA framework, we choose the recently introduced on-manifold CBF method (MCBF)~\cite{xue2024nominima}. The MCBF method applies the notion of on-manifold navigation, initially formulated for DS modulation~\cite{onManifoldMod}, to the CBF safety filter framework eliminating local minima for any arbitrarily shaped obstacle. The success of this method is attributed to escape velocities on tangent hyperplanes along the obstacle boundaries when saddle points arise.

\begin{figure*}
    \centering
    \includegraphics[width=\textwidth]{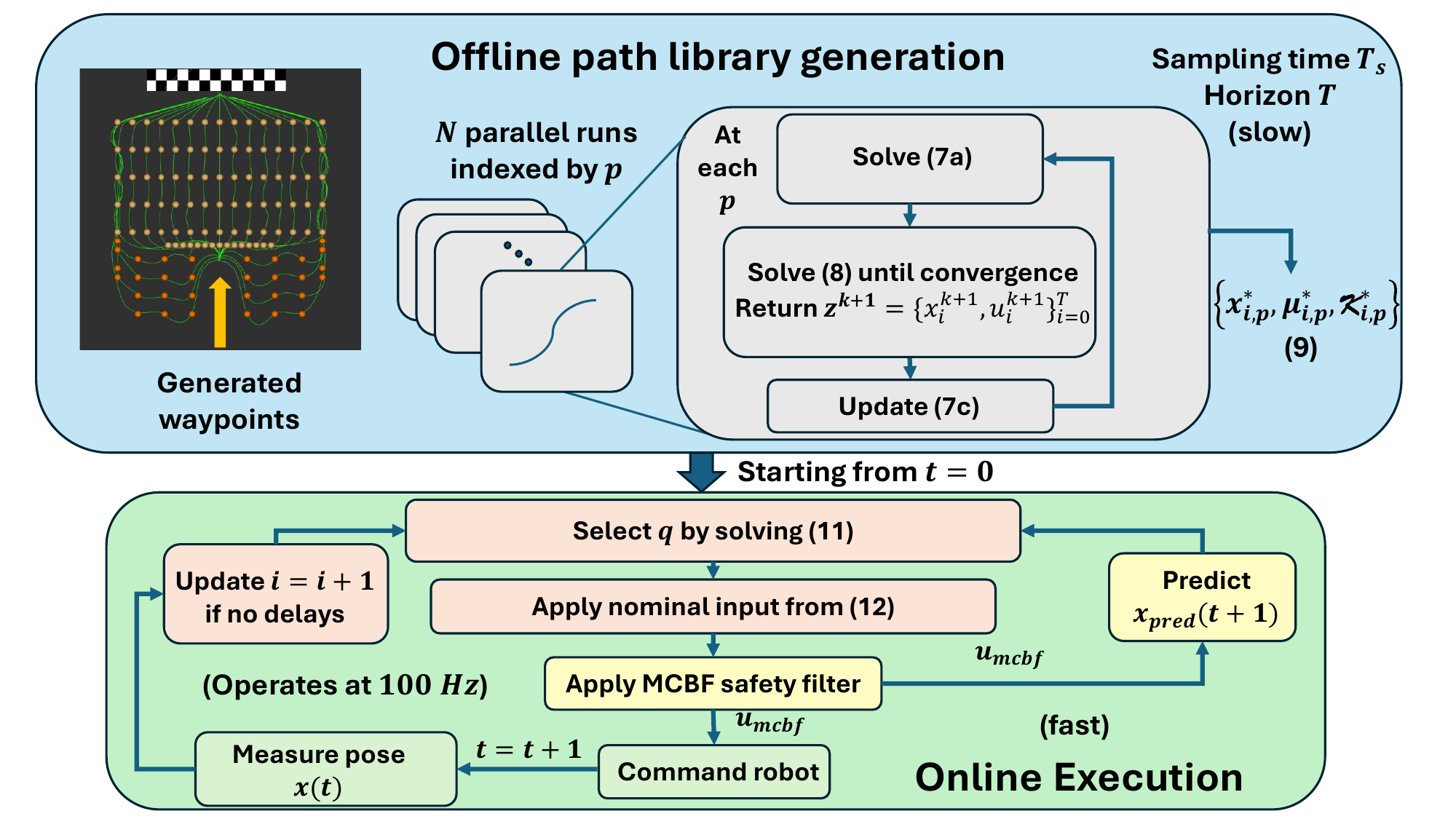}
    \caption{We present a schematic of our approach showing the offline path library generation on the top in blue and online execution in green. We generate a total of $N$ nominal controllers by using a fixed set of waypoints to sufficiently cover the initial known safe set. When commanding the robot, we use nearest neighbor search over predicted next states to choose a path where ties are broken by choosing the one closer to the center. We construct our nominal feedback input and apply a safety filter using a on-Manifold control barrier function. Finally, this is applied to the robot in real-time. 
    }
    \label{fig:intro-block-diag}
\end{figure*}

Our contributions are as follows.
\begin{enumerate}
    \item To achieve real-time performance while ensuring task completion (e.g., reaching a goal), we propose a layered solution that consists of offline path generation and an online safety filter, as shown in Figure~\ref{fig:intro-block-diag}.
    \item Using the generated library of feasible controllers, feedback gain matrices, and reference trajectories satisfying state constraints, we construct a nominal feedback law that is easy to track for reactive controllers \rev{while also ensuring smooth transitions between paths.}
    \item With online path selection and a MCBF safety filter that eliminates local minima, we ensure safety in cluttered environments with \rev{arbitrarily shaped static and dynamic obstacles.} Our online execution layer runs at $100$ Hz.
    \item We provide an open-source implementation of our approach and baselines~\footnote{\url{https://github.com/yifanxueseas/admm_mcbf_lca}}.
\end{enumerate}

In Section~\ref{sec:related-work}, we discuss prior work from related topics to contextualize our contributions, and introduce the problem in Section~\ref{sec:prob-form}. In Section~\ref{sec:admm-cbf-lca}, we propose our layered framework as shown in Figure~\ref{fig:intro-block-diag} consisting of an offline library generation layer and an online execution layer. In Section~\ref{sec:experiments}, we show comparisons of our method against baselines, highlighting the benefits of layering and long-horizon multi-path offline libraries. We end with concluding remarks in Section~\ref{sec:conclusion}.

\section{Related Work}\label{sec:related-work}
We discuss related work on two degree of freedom (2DOF) architectures and LCAs to motivate the use of the alternating direction method of multipliers (ADMM) algorithm for path generation. We also provide an overview of reactive methods for obstacle avoidance to motivate the use of MCBF. 

\subsection{2DOF and Layered Control Architectures} 
In linear control systems, the term 2DOF controllers refer to control laws that consist of a feedforward term, which drives the system to the desired trajectory, and a feedback term to compensate for errors~\cite{murray2009optimization}.  Analogous design patterns are observed in robust MPC.  For example, tube-based MPC approaches broadly apply a control input of the form $u = K(x-x_d) + u_d$, where $(x_d,u_d)$ are nominal state and control inputs computed by solving an optimization problem online, and $K(x-x_d)$ is a feedback term compensating for errors between the actual system state $x$ and the reference state $x_d$. For nonlinear systems, trajectory generation and feedback control are typically decoupled, although approaches exist that do not explicitly make this separation, e.g., iLQR~\cite{todorov2005generalized}.  
In this paper, we exploit the structure of ADMM applied to optimal control problems (OCP)~\cite{srikanthan2023augmented} to obtain a 2DOF controller with a feedforward term and feedback gains. We use the time-varying feedback gain matrices online to locally stabilize the system whenever \rev{the robot switches from one path to the next due to the presence of an obstacle}. 

\subsection{Reactive methods for Obstacle Avoidance}
Collision-free guarantees in dynamic environments are achieved either through the satisfaction of constraints in optimization-based approaches \cite{ames2019control,mpcdc2021safety,9653152,9319250,8967981,10380695} or through reactive potential-field inspired closed-form solutions \cite{apf1989real,navigationf1992exact,harmonic1997real,khansari2012dynamical,6907685,LukesDS,billard2022learning,10164805}.  Among all methods, only CBF-QP~\cite{ames2019control} can minimize deviations from nominal controller behavior while ensuring safety for general control-affine systems \rev{and a general class of arbitrary geometries~\cite{blanchini2015invariant}}. While methods to enforce robot input constraints optimally for Dynamical System motion policy modulation (Mod-DS) are proposed in \cite{xue2024nominima}, its applications are limited to fully actuated systems. Although variants of CBF-QP, such as CBF-CLF-QP are available for safe reactive navigation tasks without the use of a nominal controller, designing the Control Lyapunov Function (CLF) for navigation tasks in general is challenging and not the focus of this work. In this paper, we generate feasible paths offline to alleviate the need for enforcing tight input constraints online and guide the robot to the goal while maintaining safety through reactive methods.

\section{Background and Problem Formulation}\label{sec:prob-form}
Consider a continuous-time control-affine nonlinear system of the form 
\begin{equation}\label{eq:ct-sys}
    \dot{x}(t) = f(x(t)) + g(x(t))u(t)
\end{equation}
where $t, t_f \in \R$ are continuous-time parameters with $t$ ranging from $0$ to $t_f$, $x(t) \in \R^{n}$ is a state vector, $u(t) \in \R^m$ is a control input vector, and $f:\R^{n} \to \R^n$, and $g:\R^{n} \to \R^{n \times m}$ define the control-affine nonlinear dynamics. Our objective is to design $u(t)$ to bring the system~\eqref{eq:ct-sys} from initial state $x(0)=\xi$ to goal state $x(t_f)=\xi_g$ while remaining safe and satisfying input constraints, i.e., $u(t) \in \calU$ where 
\begin{equation}\label{eq:input-const}
    \calU = \{u \in \R^m\ |\ u_{\min} \leq u \leq u_{\max}\}.
\end{equation}

We define the notion of robot safety and consider safe control algorithms based on boundary functions. \rev{Given a twice-differentiable function $h_j$ for each obstacle $j \in \calO_{x, t}$ where $\calO_{x, t}$ is the set indices of visible obstacles if the robot is placed at state $x$ at time $t$, we define a safe set with boundary function $h_j:\R^{n} \times \R^{n} \rightarrow \R$ as the following, where $x_\text{o}^j$ is the obstacle state including its position and orientation. } %I remove the time-variant terms and replace them with obstacle states xo to keep it consistent with what we have in the appendix.
% we define a time-varying safe set with boundary function $h_j:\R \times \R^p \times \M^1 \rightarrow \R$ as
% \begin{equation}~\label{eq:safe-region}
%     \calC_t = \{x \in \R^{n_1} \times \M^{n_2}\ |\ h_j(t, x) > 0, \ \forall j \in \calO_t\}.
% \end{equation}

\rev{\begin{equation}~\label{eq:safe-region}
    \calC = \{x \in \R^{n} \  |\ h_j(x, x_\text{o}^j) > 0, \ \forall j \in \calO_{x,t}\}.
\end{equation}}

\rev{When obstacles are static, the safe set $\calC$ is fully determined by the robot state and a local field of view with visible obstacles. In this case, the index sets are only a function of robot states ($\calO_x$). In the case when obstacles are moving, we use the knowledge of obstacle velocities as well to determine $\calO_{x, t}$. We abuse notation here to define the visible obstacle indices ($\calO_{x, t}$), however, note that they are not explicitly dependent on time. They are determined by position, orientation and obstacle velocities. In future work, we aim to provide a rigorous characterization of safe sets for moving obstacles with arbitrary geometries.}

Given the above definitions, we formulate an optimal control problem (OCP) with an objective function $J$ as
\begin{equation}\label{prob:ct-ocp}
    \begin{array}{rl}
         \mathrm{minimize} &\int_{0}^{t_f} J(x(t), u(t))\\
        \mathrm{subject\ to } & \dot{x}(t) = f(x(t)) + g(x(t))u(t), \ t \geq 0, \\
        &u(t) \in \calU, \ t \geq 0, \\
        % & h_j(t, x(t)) \leq 0, \ \forall j \in \calO_t
        & \rev{h_j(x(t), x_\text{o}^j(t)) \leq 0, \ \forall j \in \calO_{x, t}}
        % & x(t) \in C, 
    \end{array}
\end{equation}
while satisfying boundary conditions given by $x(0) = \xi$ and $x(t_f) = \xi_g$. The problem in~\eqref{prob:ct-ocp} poses three challenges: a) input constraint satisfaction, b) safety, and c) real-time computation. To handle safety and real-time computation, prior work~\cite{ames2019control, loizou2007mixed, loizou2003closed} design reactive controllers to maintain the state within a time-invariant safe set. 

\rev{Recent work~\cite{mpcdc2021safety, liu2024safety}} tackles input constraint satisfaction and safety by using MPC to solve a discrete-time version of~\eqref{prob:ct-ocp}. In practice, nonlinear MPC suffers from computational burden and solver infeasibility for long-horizon problems. 

Several works~\cite{rosolia2022unified, grandia2021multi} attempts to tackle all three challenges through LCAs running low-frequency MPC without safety constraints at the higher layer and CBF-QP at the lower layer. However, without careful coordination between the layers, we empirically observe that such approaches i) still suffer from local minima in cluttered environments with concave obstacles and ii) may lead to infeasible solutions for control-affine systems with tight input constraints ($u(t) \in \calU$).

\emph{Problem Statement}: To address the combined problem of input feasibility, safety and real-time computation in~\eqref{prob:ct-ocp}, we consider the design of offline long-horizon path planners, accompanied by fast online feedback solvers and an appropriate reactive strategy to ensure safety and task completion.

\section{ADMM-MCBF as a Layered Control Architecture}
\label{sec:admm-cbf-lca}

\begin{figure}[t]
    \centering
    \begin{subfigure}{0.49\linewidth}
         \includegraphics[width=\textwidth]{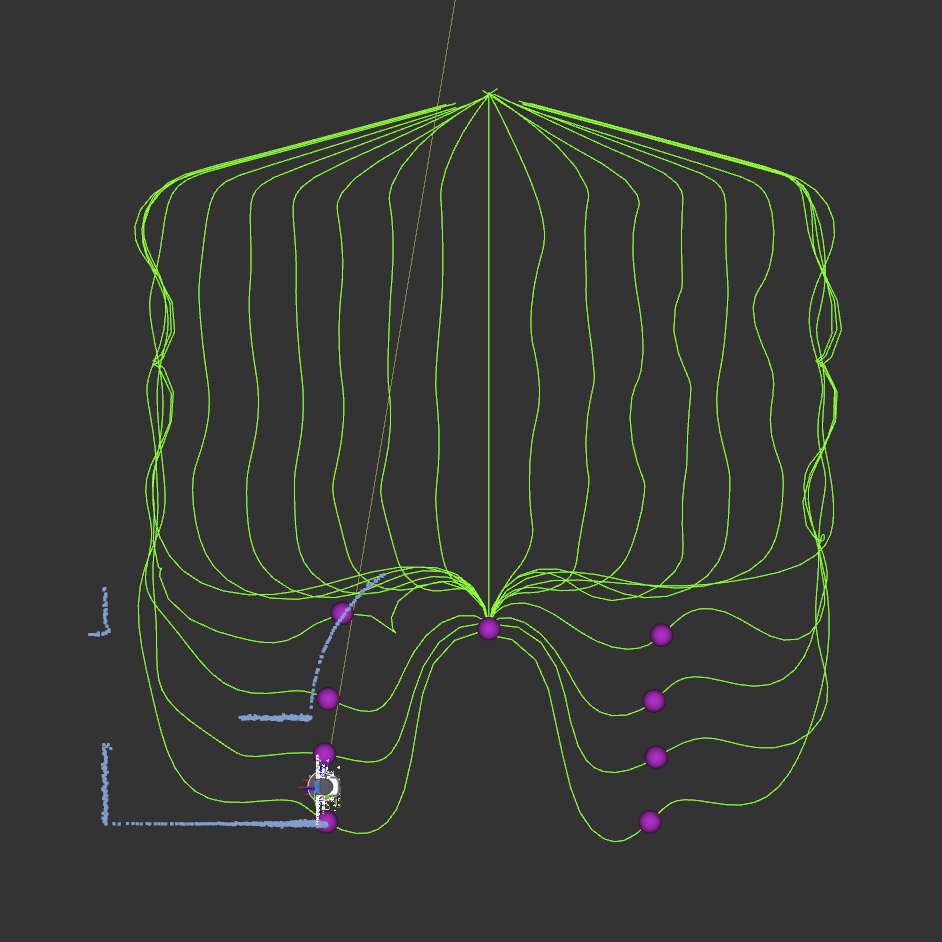}
    \caption{\label{fig:multi-path-admm}}
    \end{subfigure}
    \begin{subfigure}{0.49\linewidth}
         \includegraphics[width=\textwidth]{img/exploration space.png}
     \caption{\label{fig:hospital}}
    \end{subfigure}
    \caption{\label{fig:online-execution}In~\ref{fig:multi-path-admm}, we show the feasible trajectories from our path library in green visualizing the computed offline solutions for the initially unknown environment in~\ref{fig:hospital}. The pink dots indicate the feasible next states to choose from. The nearby detected obstacle boundaries are visualized as blue point clouds. }   
\end{figure}

Our goal is to design a long-horizon path planner to guide the system to a goal state, while leveraging reactive policies to ensure safety. To this end, we propose a layered solution that consists of an offline path planner used to generate a feasible path library, an online path selector, and an online safety filter as shown in Figure~\ref{fig:intro-block-diag}. For path library generation, we apply recent results from~\cite{o2013splitting, sindhwani2017sequential, srikanthan2023augmented} using the alternating direction method of multipliers (ADMM) algorithm to efficiently solve constrained OCPs. This provides controllers, trajectories, and time-varying feedback gain matrices to stabilize the system around the trajectory. \rev{The feedback gain matrices associated with the locally optimal trajectories play the role of a smooth proportional control action whenever the robot transitions between paths. A transition between paths is initiated whenever the robot is close to an obstacle due to the gradient velocities from the obstacle boundary function. A transition between paths may also occur due to the tangential velocities from the on-manifold constraint. In this paper, we empirically illustrate the tradeoff between incorporating constraints or lack thereof at different layers.} The trajectories in the path library are optimized by carefully sampling waypoints, as shown in Figure~\ref{fig:multi-path-admm} and described in the next subsection, from the initial location to the goal. To perform path selection online, we use a nearest neighbor search over predicted next states from the collection of paths. Once the path is selected, we compute the nominal feedback control input and solve a quadratic program with a MCBF constraint to obtain a safe input that minimally deviates from the nominal input.

\subsection{Offline Feasible Path Library Generation}
\subsubsection{Waypoint generation}
The trajectories associated with each path in the path library are regulated by two user-design choices: the inter-trajectory intervals $\delta$, and the number of waypoints per trajectory $\tau$. We assume that the users have access to the approximate size of the exploration field, i.e., how large an area the robot explores to complete a task (e.g., reaching a goal). For example, given a room of width $L$, we generate $N=L/\delta$ offline paths. We choose $\delta$ sufficiently small to provide the robot with the flexibility to avoid getting trapped. 

For most of the scenarios we consider, generating waypoints in the rectangle of dimensions $L$ by $N\delta$ between $\xi$ and $\xi_g$ is sufficient (colored yellow in \autoref{fig:intro-block-diag}). In cases where the robot travels straight towards the target, waypoints can be automatically collected from automatic grid sampling, in which points' distances with respect to each other along the length and the width of the grid are $\frac{D}{\tau}$ and $N$, where $D$ is the Euclidean distance between $\xi$'s position and $\xi_g$'s poistion. 

In rare cases where the robot is surrounded by obstacles and needs to first move away from the target, we perform a similar grip sampling algorithm but manually modify some point locations to help offline path generation algorithms converge, as shown in orange points in \autoref{fig:intro-block-diag}. 
% In rare cases when the initial state is inside the exploration space and not at a boundary, we also manually generate points colored in orange as shown in \autoref{fig:intro-block-diag}.

% We present a pseudo-code in Algorithm~\ref{alg:waypoint-generation} for waypoint generation, where $W$ is the set of all waypoints, and $w_p(k)$ denotes the $k$th waypoint associated with path $p$. The vector $\vec{d}$ connects $\xi$ to $\xi_g$, and $\vec{n}$ is normal to $\vec{d}$. 

% \begin{algorithm}
% \footnotesize
% \caption{Waypoint Generation}\label{alg:waypoint-generation}
% \begin{algorithmic}
% \Inputs{Given initial state, goal, and algorithm parameters ($\xi, \xi_g, \delta,N, \tau$)}
% \Initialize{An empty set of waypoints ($W \gets \emptyset$)}
% % \State{$W_1, W_2,\ldots,W_N \gets \emptyset$}
% \State Define the vector normal to the direction connecting initial state and goal ($\Vec{d}= \xi_g-\xi,\quad \Vec{n}\perp \vec{d}$)
% \State Define the midpoint index ($\text{mid} \gets \text{int}(\frac{N+1}{2})$)
% \For {$k=1,2,\ldots, \tau$}
% \State $w_\text{mid}(k) \gets \xi_0+\frac{k\Vec{d}}{\tau+1}$
% % \State $W_\text{mid}.\text{insert}(w_\text{mid}(k)) $
% \For {$j=1,2,\ldots, \text{mid}$}
%     \State Update $w_{\text{mid}+j}(k) = w_\text{mid}(k)+\delta j \vec{n}$
%     \State $w_{\text{mid}-j}(k) = w_\text{mid}(k)-\delta j \vec{n}$
% \EndFor
% \EndFor
% \State \Return $W =\{\{w_p(k)\}_{k=1}^{\tau}\}_{p=1}^{N} $
% \end{algorithmic}
% \end{algorithm}

\subsubsection{Nominal Controller Design}
\label{sec: offline nom design}
To design $N$ input feasible feedback controllers such that their resulting trajectories satisfy the generated waypoints, we consider a discrete-time nonlinear system obtained by the Runge-Kutta discretization of~\eqref{eq:ct-sys} with a sampling time of $T_s$ seconds as
\begin{equation}\label{eq:nonlin-sys}
    x_{i+1} = f_d(x_i, u_i)
\end{equation}
where $f_d$ is the discretized dynamics, $x_i \in \R^n$ is a state vector and $u_i \in \R^m$ is a control input vector at discrete time step $i$. Our objective is to design discrete-time feedback controllers over a finite horizon $T$ such that the input satisfies saturation constraints given by $u_i \in \calU$. We do not consider safety constraints in this design step.

Let $T/(\tau+1) = \Delta$ such that $\Delta$ is divisible by $T_s$ and the time indices at which waypoint constraints are enforced be $\calT = \{\Delta, 2\Delta, \cdots, \tau \Delta\}$. The problem is succinctly described as the following constrained OCP solved for each nominal controller indexed by $p$ as
\begin{equation}\label{prob:library-ocp}
    \begin{array}{rl}
         \mathrm{minimize} &\sum_{i=0}^{T-1} J_i(x_{i, p}, u_{i, p})\\
        \mathrm{subject\ to } & x_{i+1, p} = f_d(x_{i, p}, u_{i, p}), u_{i, p} \in \calU, \ i \geq 0, \\
        & x_{0, p} = \xi, \ x_{T} = \xi_g, \\
        & x_{i, p} = w_p(\lf i/\Delta\rf), \forall i \in \calT, 
        % & r_{i, p} = x_{i, p},\ a_{i, p} = u_{i, p}
    \end{array}
\end{equation}
where $J_i$ are stage costs, and $w_p(\cdot)$ enumerates the generated waypoints for controller $p$. To solve problem~\eqref{prob:library-ocp}, we introduce redundant state and input equality constraints as in~\cite{srikanthan2023augmented} and apply the ADMM algorithm. To simplify notation, we drop references to indices $p$ and describe our algorithm for a single nominal controller. We note that computing the $N$ nominal controllers can be easily parallelized.

For each nominal controller, define a state trajectory variable as $\boldsymbol{x}$, input trajectory variable as $\boldsymbol{u}$, and the combined trajectory variable as $\boldsymbol{z} = (\boldsymbol{x}^\top, \boldsymbol{u}^\top)^\top$. We introduce a redundant trajectory variable $\boldsymbol{\bar{z}}$ and an equality constraint $\boldsymbol{z} = \boldsymbol{\bar{z}}$ to optimization in~\eqref{prob:library-ocp}. Let $\boldsymbol{J}(\boldsymbol{\bar{z}}) = \sum_{i=0}^{T-1} J_i(\bar{z}_i)$, $\psi(\boldsymbol{z}; \xi) = \sum_{i=0}^{T-1}\psi(z_i, \xi)$ be a sum of indicator functions for dynamics constraints given as
\begin{equation*}
    \psi(z_i; \xi) = \begin{cases}
        0, & \text{if } x_{i+1, p} = f_d(x_{i, p}, u_{i, p}),\ \forall i > 0 \\
        0, & \text{if } \ x_{0, p} = \xi, i = 0 \\
        \infty, & \text{otherwise},
    \end{cases}
\end{equation*}
and $\bar{\psi}(\boldsymbol{\bar{z}}; w_p) = \sum_{i=0}^{T-1} \bar{\psi}(\bar{z}_i; w_p)$ be the sum of indicator functions for waypoints and input constraints defined as
\begin{equation*}
    \bar{\psi}(\bar{z}_i; w_p) = \begin{cases}
        0, & \text{if } i \in \calT,\  \bar{z}_{u, i} \in \calU, \text{ and } \bar{z}_{x, i} = w_p(i/\Delta), \\
        0, & \text{if } i \notin \calT,\ \bar{z}_{u, i} \in \calU, \bar{z}_{x,T} = \xi_g\\
        \infty, & \text{otherwise},
    \end{cases}
\end{equation*}
where $\bar{z}_{x,i}$ and $\bar{z}_{u, i}$ correspond to state and input redundant variables at time $i$, respectively.

To obtain a layered decomposition~\cite{srikanthan2023augmented}, we apply the ADMM algorithm to~\eqref{prob:library-ocp}, yielding the following iterative updates 
\begin{subequations}\label{prob:ocp_admm}
    \begin{align}
        \boldsymbol{\bar{z}}^{k+1} &\coloneqq \argmin\limits_{\boldsymbol{\bar{z}}} \boldsymbol{J}(\boldsymbol{\bar{z}};\xi_g) + \frac{\rho}{2} \left \| \boldsymbol{z}^k - \boldsymbol{\bar{z}} + \boldsymbol{v}^k \right \|_2^2 \nonumber \\
        \text{s.t. } & \bar{\psi}(\boldsymbol{\bar{z}}; w_p) = 0, \label{eq:trajectory-gen} \\
        \boldsymbol{z}^{k+1} &\coloneqq \argmin_{\boldsymbol{z}} \ \frac{\rho}{2} \| \boldsymbol{z} - \boldsymbol{\bar{z}}^{k+1} + \boldsymbol{v}^k \| \nonumber \\
        \text{s.t. } & \psi(\boldsymbol{z}, \xi) = 0 \label{eq:fb-control} \\
        \boldsymbol{v}^{k+1} &\coloneqq \boldsymbol{v}^k + \boldsymbol{z}^{k+1} - \boldsymbol{\bar{z}}^{k+1} \label{eq:dual-update},
\end{align}
\end{subequations}
where $\boldsymbol{z, \bar{z}}$ are primal variables, $\boldsymbol{v}$ is a scaled dual variable and $\rho$ is a tuning parameter to improve convergence of the ADMM algorithm. Moreover~\eqref{eq:trajectory-gen} is a time separable feasibility problem  without dynamics constraints that can benefit from parallelization, while~\eqref{eq:fb-control} is an unconstrained OCP with nonlinear dynamics and a quadratic objective function which can be solved via sequential linearization-based techniques, such as iLQR. Finally, the dual updates in~\eqref{eq:dual-update} ensure consistency so as to guarantee both constraint satisfaction and dynamic feasibility at convergence. We note here that one could choose any constrained optimization method to solve~\eqref{prob:library-ocp}, however, using the proposed ADMM-based decomposition facilitates computing feedback gain matrices, as discussed below. 

If $\boldsymbol{J}$ is convex, the only nonconvex part of the problem is the \emph{unconstrained} nonlinear dynamics found in the update step~\eqref{eq:fb-control}. By isolating the nonconvexity of the problem to this update step, we can leverage iLQR~\cite{todorov2005generalized} to rapidly converge to a locally optimal solution under fairly benign assumptions~\cite{liao1991convergence}. In what follows, we show how we solve~\eqref{eq:fb-control} using iLQR, resulting in local convergence to an optimal feedback controller that tracks the reference trajectory $\boldsymbol{\bar{z}}$ from~\eqref{eq:trajectory-gen}. 

\subsubsection{Feedback controller from ADMM iterates}
\label{sec: nom fd controller}
For any ADMM iteration $k+1$, let $\boldsymbol{d}^k = -(\boldsymbol{z}^k - \boldsymbol{\bar{z}}^{k+1} + \boldsymbol{v}^k)$. By doing a change of variable from $\boldsymbol{x, u}$ to $\boldsymbol{\delta x} = \boldsymbol{x} - \boldsymbol{x}^k$ and $\boldsymbol{\delta u} = \boldsymbol{u} - \boldsymbol{u}^k$, and setting $\boldsymbol{\delta z} = (\boldsymbol{\delta x^\top, \delta u^\top})^\top$, the feedback control problem in~\eqref{eq:fb-control} is approximately solved via iLQR:
\begin{align}
    \boldsymbol{\delta z}^{\ell+1} \coloneqq &\argmin_{\boldsymbol{\delta z}}\ \frac{\rho}{2}\| \boldsymbol{\delta z} - \boldsymbol{d}^{\ell} \|_2^2 \nonumber \\
        &\text{s.t. } \delta z_{x,i+1} = \begin{bmatrix}
            A_i^{\ell} & B_i^{\ell} \end{bmatrix}
            \delta z_i \label{eq:nonlin-track-layer} 
\end{align}
where $\ell$ is iLQR iteration counter and the constraint in~\eqref{eq:nonlin-track-layer} is from linearization of the dynamics constraints $f_d$ with $\begin{bmatrix}
    A_i^{\ell} & B_i^{\ell}
\end{bmatrix} = \frac{\partial f_d(z)} {\partial z}|_{z=z^{k+\ell}}$. Solving~\eqref{eq:nonlin-track-layer} using dynamic programming results yields time-varying feedback gain matrices $\calK_{i, p}^\star$~\cite{todorov2005generalized}. 
Once the ADMM algorithm has converged for computing the $p$th nominal controller, we have for any discrete-time index $i$
\begin{equation}\label{eq:ctrl-params}
    \mu_{i, p}^\star, \quad \calK_{i, p}^\star, \quad x_{i, p}^\star
\end{equation} 
the nominal input, the feedback gain matrix, and the resulting state vector respectively. We next \rev{illustrate} how the feedback gain matrices from each nominal controller $\calK_{i, p}^\star$ play a role in stabilizing the system online whenever the MCBF safety filter modifies the nominal inputs.

\subsection{Online Safe Reactive Control}

In this section, we first describe the MCBF controller we solve online using the selected nominal inputs and provide details of our proposed architecture in the green block of Figure~\ref{fig:intro-block-diag}.

\subsubsection{Reactive Control with MCBF} While our proposed LCA could use any reactive strategy, to generate safe inputs that guarantee no local minima we use the modulation based on-Manifold Modulated CBF-QP (MCBF-QP) \cite{xue2024nominima}. Given certain nominal controllers $u_\text{nom}$, controllers like standard CBF-QPs lead to inevitable saddle points, where the robot stops on obstacle boundaries before it reaches the target. The proof of this for fully-actuated robots can be found in Theorem 5.3 of \cite{xue2024nominima}. To overcome the saddle points of CBF-QP solvers and their variants, MCBF-QP modulates the dynamics and projects components onto the tangent planes of the obstacle boundary functions, introducing velocities that eliminate saddle points. This is achieved by introducing an additional constraint on the standard CBF-QP formulations defined as \eqref{eq:mod-phi-cbf constraint affine}. MCBF-QP and CBF-QP can be generalized to any nonlinear affine systems, where $L_f$ and $L_g$ are the Lie derivatives of $h(x, x_o)$.
% \begin{equation}
%     \phi(x,t)^\top f(x)+\phi(x,t)^\top g(x)(u -u_\text{nom})\geq \gamma. \label{eq:mod-phi-cbf constraint affine}
% \end{equation}
\begin{gather}\label{eq:mod-phi-cbf affine} 
\nonumber u_{\text{mcbf}} = \argmin_{u \in \mathbb{R}^p}\;(u-u_\text{nom})^\top(u-u_\text{nom})\\
L_fh(x,x_o) + L_gh(x,x_o)u + \nabla_{x_o}h(x,x_o)\dot{x}_o\geq -\alpha (h(x,x_o))\\
\phi(x,x_o)^\top f(x)+\phi(x,x_o)^\top g(x)u \geq \gamma \label{eq:mod-phi-cbf constraint affine}
\end{gather}
Here, $\gamma$ is the user's choice of a strictly positive real number, and the tangent direction $\phi(x)$ can be derived using various obstacle exit strategies, such as hessian or geometric approximation. See \cite{xue2024nominima} and \cite{onManifoldMod} for implementation details. The resulting MCBF filter eliminates local minima for unsafe sets of all geometric types, including non-star-shaped obstacles, as shown in \cite{xue2024nominima}. 

\subsubsection{Online Path Selector} To achieve obstacle avoidance in real-time and ensure task completion, we initialize the path at $t=0$ using the center-line connecting initial state to the goal. For $t>0$, we first run a nearest neighbor search algorithm to decide which path to track:
\begin{equation}
    q(t+1) = \argmin_p \| \{x_{i, p}^\star\}_{p=0}^N - x_{pred}(t+1) \|,
\end{equation}
where $x_{pred}(t+1) = f_d(x(t), u_{mcbf}(t))$ with $u_{mcbf}(t)$ as the optimized input from solving MBCF-QP~\cite{xue2024nominima}. Ties are broken by selecting the one closer to the center-line connecting the initial state to the goal. Once we choose $q(\cdot)$, we construct the nominal input with a feedback term as 
\begin{equation}\label{eq:nominal-inp}
    u_{nom}(t) = \mu_{q, i}^\star + \calK_{i, q}^\star(x_{i, q}^\star - x(t)).
\end{equation}
Next, we apply the MCBF filter % solving~\eqref{eq:mod-phi-cbf affine} 
to ensure obstacle avoidance, and command the robot. The feedback control law in~\eqref{eq:nominal-inp} has a feedforward term $\mu_{q, i}$ that drives the robot toward the goal and a feedback term $\calK_{i, q}^\star$ that compensates for deviations from $x_{i, q}^\star$. We emphasize that these feedback gain matrices $\calK_{i, q}^\star$ are derived from~\eqref{prob:ocp_admm}, not tuned manually.
The key insight is in using time-varying feedback gain matrices $\calK_{i, q}^\star$ to construct a ``proportional term'' to stabilize the system when the nominal input~\eqref{eq:nominal-inp} is modified. When no obstacles are present, $u_{mcbf} \approx u_{nom}$, while in obstacle situations, $\calK_{i,q}$ compensates for the deviations. Our online execution runs at 100 Hz, solving in under 0.01 seconds on average.

\section{Experiments}\label{sec:experiments}
\begin{figure}[t]
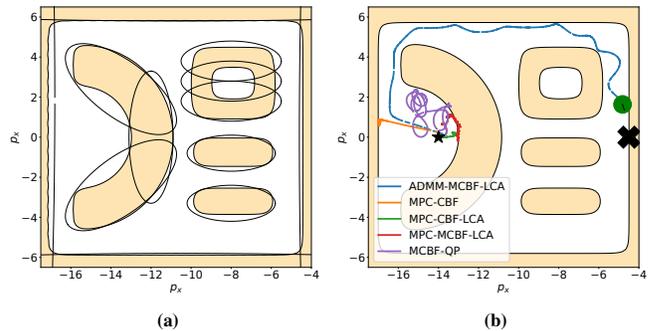

    \centering
    \begin{subfigure}{0.49\linewidth}
         \includegraphics[width = \textwidth]{img/env approximation.pdf}

    \caption{\label{fig:env approx}}
    \end{subfigure}
    \begin{subfigure}{0.49\linewidth}
         \includegraphics[width=\textwidth]{img/hospital trajectories.pdf}

     \caption{\label{fig:hospital-traj}}
    \end{subfigure}

    \caption{In~\ref{fig:env approx}, we show boundaries of the obstacle inside hospital Gazebo simulations that are learned using GPDF, and the ellipse approximations used by MPC-CBF solver. In~\ref{fig:hospital-traj}, we plot the results from Gazebo simulations of our approach and $4$ baselines for the indoor environment with unknown obstacles. We observe that ours shown in blue is the only approach that reaches the goal while ensuring safety.}   

\end{figure}

\begin{table*}
    \begin{tabular}{c|c|c|c|c|c|c|c}
    \hline
        Horizon ($T$) & Sampling Time ($T_s$) & Avg. Time (secs) \\
         \hline
        16 & 0.5 &  2.090\\
        \hline
        40 & 0.2 & 7.545 \\ 
        \hline
        80 & 0.1 &  23.175\\ 
        \hline
        160 & 0.05 & 47.635\\
        \hline
        400 & 0.02 & 209.595\\
        \hline
    \end{tabular}
    \caption{We report the time taken to compute $5$ paths for different values of horizon $T$ and $T_s$ on the unicycle model for $5$ randomly sampled initial conditions given the same goal condition. In our GitHub repository, we also provide several other dynamics models to test on.}
    
    \label{tab:admm}
\end{table*}

\begin{table*}[!tbp]
\centering
\begin{minipage}[t]{1.0\textwidth}
\begin{center}
    \centering
    \resizebox{\linewidth}{!}{\begin{tabular}{c|c|c|c|c|c|c|c}
    \hline
        Algorithms & QP-solver fail \# $\downarrow$ & Safety \% $\uparrow$ & Task success \% $\uparrow$ & Lin. speed ($\Bar{v}$) $\uparrow$ & Ang. speed ($\Bar{\omega}$) $\downarrow$ & $v$ err. ($e_v$) $\downarrow$ & $\omega$ err. ($e_\omega$) $\downarrow$ \\
         \hline
        \textit{ADMM-MCBF-LCA} & \textbf{0} &  \textbf{100}&  \textbf{100} & \textbf{0.487} & 0.584 & 0.012 & 0.027 \\
        \hline
        \textit{MPC-CBF}  & - & 100 & 0 & 0.100 & 1.303 & 0.011 & 0.074\\ 
        \hline
        \textit{MPC-CBF-LCA} & 0 & 100 & 0 & 0.048 & 0.537 & 0.005 & 0.024 \\ 
        \hline
        \textit{MPC-MCBF-LCA}  & 33 & 100 & 0 & 0.024 & 1.246 & 0.008 & 0.037 \\
        \hline
        \textit{MCBF-QP} & 19 & 60 & 0 & 0.380 & 0.989 & 0.015 & 0.051 \\ 
        \hline
        \textit{MPPI}  & - & 100 & 0 & 0.036 & 0.312 & 0.037 & 0.035\\ 
        \hline
    \end{tabular}}
    \caption{\textmd{We report the metrics for all algorithms evaluated from 5 gazebo trials showing in bold the metrics for which our proposed ADMM-MCBF-LCA performed the best. We use $\uparrow$ and $\downarrow$ to indicate if higher or lower is better, respectively. $\Bar{v}$ and $\Bar{\omega}$ are the average linear and angular speeds and their average tracking errors are denoted by $e_{\Bar{v}}, e_{\Bar{\omega}}$. The first three columns indicate algorithm performance in terms of QP solver failure across trials, \% collision-free trials, and \% task completion rate.} 
    }
    \label{tab:uni}
\end{center}
\end{minipage}
\end{table*}

In this section, we evaluate our proposed method on the differential drive robot Fetch through realistic Gazebo simulations and qualitative hardware experiments. We compare our proposed ADMM-MCBF-LCA against five baselines, each representing a category of prior work on safe real-time robot navigation in dynamic environments. We use metrics such as input tracking error, solver infeasibility, collision-free rate, and task success. For ADMM, we use the $\rho$-update rule described in~\cite[\S3.4.1, equation (3.13)]{boyd2011distributed} to improve convergence. We provide custom implementations of all algorithms using \texttt{cvxpy}~\cite{diamond2016cvxpy}, CasADi~\cite{Andersson2019} and JAX optimizers~\cite{jax2018github}. Empirically, we found $\delta=0.8 m$ to be a reasonable interval size for Fetch robot navigation tasks for our dynamic environment settings. 

\subsection{Robot Dynamics and Obstacle Definitions}\label{subsec:rob-dyn}
For all experiments, we model Fetch using the unicycle nonlinear dynamics given as 
\begin{equation}
\label{eq:standard unicycle}
    \dot{x}=\begin{bmatrix}
        \dot{p_x}\\\dot{p_y}\\\dot{\theta}
    \end{bmatrix}=
    \begin{bmatrix}
        \cos{\theta} & 0\\ \sin{\theta} & 0\\ 0 & 1
    \end{bmatrix}
    \begin{bmatrix}
        v \\ \omega,
    \end{bmatrix}
\end{equation}
where $v, \omega$ are the linear and angular speeds such that $v_{max} = 1$m/s, $\omega_{max} = 2$m/s, $v_{min} = \omega_{min} = 0$. The state of the robot consists of 2-D Euclidean position $p_x, p_y$, and orientation $\theta$.

To apply the MCBF-QP and CBF-QP approaches, the CBF must have relative degree 1, i.e., the control input must appear in the first time-derivative of the CBF. Since the control input $\omega$ for~\eqref{eq:standard unicycle} is of relative degree 2 to the CBFs used, instead of a signed distance function we use a higher-order CBF
(HO-CBF)~\cite{diffBarrier, diffCBF} of relative degree $1$ with respect to control inputs $v, \omega$. Our experiments highlight how judiciously combining a properly designed global nominal controller with a MCBF-QP filter (based on a HO-CBF) can achieve stable and safe navigation in crowded and concave environments.  For all methods except MPC-CBF, the obstacle boundaries are obtained using learned Gaussian Process Distance Fields (GPDF)~\cite{choi2024towards}. Since MPC-CBF requires discrete-time CBF constraints, we instead use ellipse approximations, as shown in~\autoref{fig:env approx}. In our approach, the index $i$ is updated based on progress to $x_{i,q}^\star$ using distance measurements, adjusting for delays within a user-specified tolerance.

\subsection{Results}

The task for the robot is to navigate an unknown indoor hospital environment as shown in Figure~\ref{fig:hospital-traj} starting within the concave obstacle desk and reach the other end of the room. Among all approaches, only our approach safely navigates the robot to the goal shown in Figure~\ref{fig:hospital-traj}. We summarize metrics in~\autoref{tab:uni}, where $\Bar{v}$, $\Bar{\omega}$, are average linear and angular speeds, $e_{\Bar{v}}, e_{\Bar{\omega}}$ are errors in tracking linear and angular speeds, computed as deviations between commanded and measured inputs. On hardware as shown in Figure~\ref{fig:experiment}, we show our robot safely navigating around a human walking in the robot's exploration space. 

From the results reported, we observe that MCBF-QP controller fails to reach the goal although it is guaranteed to escape local minima. We believe that this is due to large model mismatch from using a naive dynamics model for nominal input design. Using discrete-time MPC with CBF constraints (MPC-CBF) struggles with navigating Fetch out of concave obstacles due to large tracking errors from low update frequencies. The nonlinear CBF constraints increase runtime, limiting the horizon size; with a horizon of 5, CasADi solvers drop below 5Hz. Additionally, MPC-CBF often fails to find globally optimal solutions, returning infeasible across all trials. As reported in the table~\ref{tab:uni}, MPC-CBF results in large average angular speeds and input tracking errors. \change{The sampling-based variant of MPC, biased Model Predictive Path Integral (MPPI) in~\cite{biasedMPPI}, demonstrates the same issues as MPC-CBF-LCA, resulting in local minima at the concave obstacle region preventing the robot from finding a way out.} 
Among the layered baselines, we find that MPC with CBF-QP outperforms MPC with MCBF-QP resulting in lower input errors and lower average $\Bar{\omega}$. However, neither approaches escape the concave obstacle region to reach the goal. Through the careful design of a nominal feedback law as in~\eqref{eq:nominal-inp}, our proposed ADMM-MCBF-LCA successfully navigates the robot out of the concave obstacle to the goal. In supplementary videos, we show real-robot experiments of our approach outperforming all baselines in indoor environments with static and dynamic obstacles.

\section{Conclusion} \label{sec:conclusion}
We conclude that addressing the combined challenges of input feasibility, safety, and real-time computation for robot navigation in environments with moving obstacles, requires layered solutions for robustness and task performance. Our proposed ADMM-MBCF-LCA generated a library of feasible paths offline, to construct a nominal feedback law that is easy to track for our MCBF filter to ensure safety. We demonstrated experiments on realistic Gazebo simulations and hardware showing our approach outperforming baselines in terms of reaching the goal while generating safe and feasible inputs. The baselines representing prior work on reactive methods, and end-to-end solvers perform worse than all layered approaches. We also observed that due to the computation-performance tradeoffs, nonlinear MPC based LCAs do not reach the goal and are stuck in local minima. In future work, we plan to address the question of ADMM convergence rigorously for nonlinear systems. 
\bibliographystyle{abbrvnat}

\bibliography{papers}

\appendix

\section{Details of barrier methods for collision avoidance}

For example, let us first consider a static circular obstacle with a boundary function of the form
\begin{equation}\label{eq:circle}
    h(x) = \|x - x_c \|_2^2 - c_r^2
\end{equation}
where $x_c, c_r$ are the center and radius of the circle, and $x \in \R^2$ is the 2-D position. We have that when $h(x) > 0$ the robot lies outside the obstacle boundary, $h(x) = 0$ means that the robot is on the obstacle boundary, and $h(x) < 0$ implies that the robot is inside the obstacle. The safe set associated with $h$ is given by
\begin{equation}\label{eq:safe-set}
    \calC = \{x \in \R^2\ |\ h(x) > 0\}.
\end{equation}

\begin{definition}
    A safe set $\calC$ is positively-invariant if for every initial state $x(0) \in \calC$, we have that $x(t) \in \calC$ for all $t > 0$.
\end{definition}

\begin{definition}
    A scalar continuous function $\alpha(r)$, belongs to class $\calK_{\infty}$ if it is strictly increasing such that $\alpha(0) = 0$ for $r = 0$ and $\alpha \rightarrow \infty$ as $r \rightarrow \infty$.
\end{definition}

\begin{definition}
    The continuously differentiable function $h$ is a control barrier function on the safe set $\calC$ as defined in~\eqref{eq:safe-set}, if there exists a class $\calK_{\infty}$ function that satisfies
    \begin{equation*}
        \sup_{u \in \calU} \nabla_x h(x)^T \dot{x} \geq \alpha(h(x)), \forall x \in \calX.
    \end{equation*}
\end{definition}
% Following the results from~\cite{singletary2021comparative} connecting the literature on artificial potential fields and control barrier functions, we consider a derivative condition on $h$ as
% \begin{equation*}
%     \dot{h}(x) = \nabla_x h(x)^T \dot{x} \geq -\alpha(h(x)).
% \end{equation*}

Continuing our example for the static circular obstacle, we calculate the derivative to obtain
\begin{equation*}
    2(x - x_c) v \geq \alpha(h(x)).
\end{equation*}

\begin{lemma}
    For the dynamical system $\dot{x} = v$, let $h$ as defined in~\eqref{eq:circle} be a control barrier function for a safe set $\calC$ defined in~\eqref{eq:safe-set}. Now, let $v \in K(x)$ be such that $K(x) = \{v \in \calU\ |\ \nabla_x h(x)^T v \geq \alpha(h(x))\}$ is a locally Lipschitz feedback control law. Then, for all $x(0) \in \calC$ we have that $x(t) \in \calC$ for $t \in [0, T)$. The proof follows from~\cite[Definition 4.9]{blanchini2015invariant}.
\end{lemma}

% \begin{proof}
%     For a given set $\calC$ to be a practical set, it needs to satisfy the following conditions. 
%     \begin{enumerate}
%         \item The safe set $\calC$ defined in 
%     \end{enumerate}
%     % Check if the set $\calC$ is a practical set. Meaning check if it satisfies the conditions defined in~\cite[Definition 4.9]{blanchini2015invariant}. Need to write proof.
% \end{proof}

% Will this hold even if $v = u_{nom} + K(x - r)$ as in our setting which is a discrete-time solution? Need to check if the resulting nominal feedback control law is locally Lipschitz in $x$. Maybe this is true because of the use of iLQR gains? Need more intuition on this. This is not obvious to show. This is not the focus of this paper though.

We extend our analysis for the robot dynamics from Section~\ref{subsec:rob-dyn} by introducing a higher order control barrier function given by
\begin{equation}
    \bar{h}(x, \theta) = h(x) + \langle w_h\nabla_x h(x), \vec{b}\rangle,
\end{equation}
where $\vec{b} = (\cos{\theta}, \sin{\theta})$ and $w_h$ is a user-defined constant. We note here that we require the boundary function $h$ to be twice-differentiable to define the derivative condition on $\bar{h}$. The safe set associated with $\bar{h}$ is defined on $(x, \theta)$.

The derivative condition on $\bar{h}$ is given by
\begin{equation*}
    \dot{\bar{h}}(x, \theta) = \nabla_x h(x) v + \langle w_h \nabla^2_x h(x), \vec{b} \rangle + \langle w_h \nabla_x h(x), \nabla_{\theta} \vec{b} \rangle \geq \alpha(\bar{h}(x, \theta).
\end{equation*}
We provide the computation of the derivative using JAX libraries in our GitHub repository~\footnote{\url{https://github.com/yifanxueseas/admm_mcbf_lca}}. 

Now, if the circular obstacle was moving, i.e., say the circular obstacle is dynamic. Then, the boundary function associated with the moving obstacle is given by
\begin{equation*}
    h(x, x_c) = \|x - x_c \| - c_r^2,
\end{equation*}
where $x_c$ has its own dynamics as the obstacle moves.

We calculate the derivative associated with the dynamic obstacle as
\begin{equation*}
    \dot{h}(x, x_c) = \frac{\partial h}{\partial x} \dot{x} + \frac{\partial h}{\partial x_c} \dot{x_c} = 2(x - x_c) v - 2(x - x_c)v_{obs} \geq \alpha(h(x)),
\end{equation*}
where $v_{obs}$ is the velocity of the obstacle. Instead of using a motion prediction model, we use a finite-differencing method to compute the velocities of obstacles from real-time measurements. The higher order control barrier function associated with the moving obstacle for the unicycle robot dynamics now becomes 
\begin{equation*}
    \bar{h}(x, \theta, x_c) = h(x, x_c) + \langle w_h \nabla_x h(x, x_c), \vec{b} \rangle,
\end{equation*}
where $\bar{h}$ is a function of both robot and obstacle states. The safe set associated with the boundary function in the dynamic obstacle setting is defined as a level set of $(x, \theta, x_c)$. 

For obstacles that aren't circular, we define a boundary function using the tool from~\cite{choi2024towards} that fits a Gaussian Process (GP) based distance function using point cloud measurements from the surfaces of obstacle boundaries. The analysis for checking if the safe set satisfies the conditions of a practical set for the GP-based distance function is left as future work.

\end{document}